\documentclass[letterpaper, 10 pt, conference]{ieeeconf}  
\usepackage{xcolor}

\IEEEoverridecommandlockouts                              

\usepackage{graphicx} 
\usepackage{mathptmx} 
\usepackage{amsmath} 
\usepackage{amssymb}  
\usepackage{cite}
\usepackage{stfloats}
\usepackage[linesnumbered,ruled,vlined]{algorithm2e}
\usepackage{eucal}

\title{\LARGE \bf
A Vision-Based Collision Sensing Method for Stable Circular Object Grasping with A Soft Gripper System
}

\author{Boyang Zhang$^{1}$, Jiahui Zuo$^{1}$, Zeyu Duan$^{2}$, and Fumin Zhang$^{1}$, \textit{Fellow, IEEE} 
\thanks{*This work was not supported by any organization}
\thanks{$^{1}$B. Zhang, J. Zuo, and F. Zhang (corresponding author) are with the Department of Electronic and Computer Engineering, The Hong Kong University of Science and Technology, Hong Kong (email: bzhangcd@connect.ust.hk, jzuoai@connect.ust.hk, eefumin@ust.hk).}%
\thanks{$^{2}$Z. Duan is with the Department of Civil and Environmental Engineering, The Hong Kong University of Science and Technology, Hong Kong (zy.duan@connect.ust.hk).}%
}

\begin{document}
\setlength{\textfloatsep}{5pt}

\maketitle
\thispagestyle{empty}
\pagestyle{empty}

\begin{abstract}

External collisions to robot actuators typically pose risks to grasping circular objects. 
This work presents a vision-based sensing module capable of detecting collisions to maintain stable grasping with a soft gripper system.
The system employs an eye-in-palm camera with a broad field of view to simultaneously monitor the motion of fingers and the grasped object. 
Furthermore, we have developed a collision-rich grasping strategy to ensure the stability and security of the entire dynamic grasping process. 
A physical soft gripper was manufactured and affixed to a collaborative robotic arm to evaluate the performance of the collision detection mechanism. 
An experiment regarding testing the response time of the mechanism confirmed the system has the capability to react to the collision instantaneously.
A dodging test was conducted to demonstrate the gripper can detect the direction and scale of external collisions precisely.


\end{abstract}

\section{Introduction}

Collisions widely exist in everyday life manipulation tasks, which impact various aspects of the interactions between humans and objects from industrial to household scenarios. 
Generally, the collision has both negative and positive sides. 
On the negative side, it poses a safety hazard to make humans hurt or damage the objects.
On the opposite, the collision can also provide an effective signal for perceiving the contact with objects. 
Humans have a complete and sophisticated sensory system to perceive collision information when executing a manipulation task.
While for a robotic system, it also requires collision sensing to improve the stability of tasking with manipulating objects, such as stacking plates, picking up harvested fruits, and twisting bottle caps.
Furthermore, the positive side of collision can be utilized as sensitive contact information in a grasping strategy. 
When humans try to grasp something, they always contact objects first to assist in localizing the object before the action of grasping. 
Inspired by humans, a collision-based grasping strategy is proposed to help the robot localize the objects without explicit depth data.

The manipulation of circular objects often increases the difficulty for humans and robots.
The geometric properties of circular objects introduce the challenges: limited contact area and curvature surface.
Therefore, a soft gripper system is designed for the verification of collision sensing rather than a traditional gripper.
The gripper has three pneumatic soft fingers to increase the area and adapt to the slippery surface.
It integrates a camera module at the center of the gripper's palm, providing real-time fisheye vision.
This design leverages the fisheye camera to monitor objects and finger states simultaneously through its large field of view (FOV).

\begin{figure}[t]
    \centering
    \includegraphics[width=1.0\linewidth]{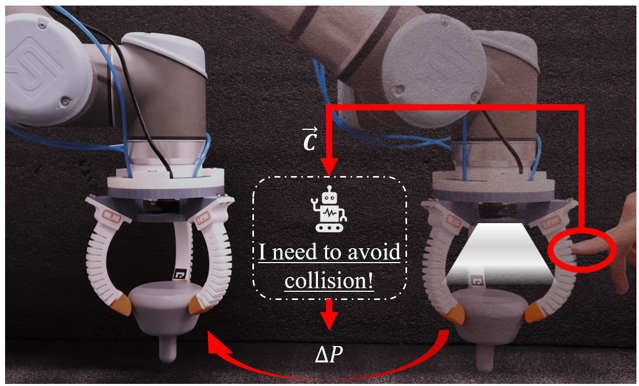}
    \caption{An explicit vision-based collision sensing can be a component in the perception and decision of robots. }
    \label{init}
\end{figure}

An experimental campaign has been conducted to verify the sensitivity of the gripper system with clear visualization. It illustrates the response time of the gripper is close to the reaction of humans' tactile sensing and can be adaptive to merge into the control framework of the robot arm. Based on this characteristic, the problem in this work can be formulated as: (1) How to detect reliable collision information? (2) How to establish the relationship between collision and action of the robot?
The precise direction and scale of vector $\overrightarrow{C}$, which represents collision information, can lead the robot to leave the collision area on the correct path. 
Under the coordinate of the robot, the relationship will be indicated as the variation of position $\Delta{P}$, as shown in Fig. \ref{init}.

In summary, key contributions of this work include:

\begin{itemize}
\item During grasping, a collision-based grasping strategy is proposed to achieve stable localization in a long-term grasping process based on the integration of a practical gripper system mounted on the robot arm.
\item At the post-grasping period, a vision-based real-time collision detection mechanism with pixel-wise deformation analysis is developed to protect the grasping process from potential external disturbances.
\end{itemize}


\section{Related Work}

\subsection{Collision in Manipulation}
There are two directions related to collisions in manipulation: collision-based (collision-rich) manipulation and collision avoidance. 
Collision-based manipulation means there are various formats of contact appearing in the process of manipulation, which act as a significant participant in the integration of the sensing module. 
\cite{9217019} proposed a tactile-based robot arm that sensitively perceives the external force and realizes safe human-robot interaction. 
In \cite{10706767}, tactile-based grasping was demonstrated, which incorporates light contact detection, grasp pose adjustment, and loss-of-contact detection to enhance the reliability of object grasping. 
\cite{10138678} proposed a new soft hand with haptic sensing to realize object classification and precise grasping adjustments. 
In \cite{9839393}, the researchers used a binary collision signal to lead the robot to make the correct decision in orientation. 

Collision avoidance is crucial for ensuring robust and stable grasping.
Traditional solutions usually utilize global observation systems with cameras to reconstruct the entire environment, requiring significant computing resources and complex algorithms like \cite{10161529}.
While these systems excel at macro-level collision avoidance, they lack the real-time responsiveness necessary for precision tasks. 
To overcome these limitations, sensing modules are integrated into the gripper to offer solutions. \cite{1087328} utilized several infrared proximity sensors on the gripper fingers to perceive the surrounding environment.
\cite{9144379} integrated a multi-sensory system on the gripper to create a sensory model for accurately gathering distance information to prevent collisions with the table.
\cite{10144527} proposed a hybrid vision and tactile sensing-based collision detection method.

\subsection{Circular Object Grasping}
According to the Cutkosky taxonomy \cite{34763}, the grasping model in this work is classified as precise, compact, and circular, focusing on sensitivity. 
However, this model prioritizes geometric fit over stability and security, which is a key aspect addressed in our work.
Previous research in circular object grasping has concentrated on theoretical physical modeling and analysis. 
For instance, works such as \cite{7354266} optimized dynamic grasping methods based on soft fingertips, while \cite{4398964} proposed a friction-dependent grasp posture for circular objects. 
In \cite{5723563}, researchers principally analyzed the geometric properties of grasped circular objects. 
In addition, innovations in structural design have been explored, as demonstrated in \cite{4059336} with the design of multi-DOF robotic fingers.
Further research has focused on enhancing gripper design for achieving more stable grasping of circular objects.
\cite{10341385} introduced a 3D-printed adaptive gripper with a unique fluff-like gripping surface structure. 
\cite{10122008} presented a donut-shaped soft gripper designed for secure grasping of circular objects. 
Drawing from the advantages of these works, our research integrates both systematic analyses and practical experiments.



\section{Design of Gripper System}
\label{sec3}

\subsection{Appearance and Features} 
\label{sec3_1}

\begin{figure}[h]
    \centering
    \includegraphics[width=0.92\linewidth]{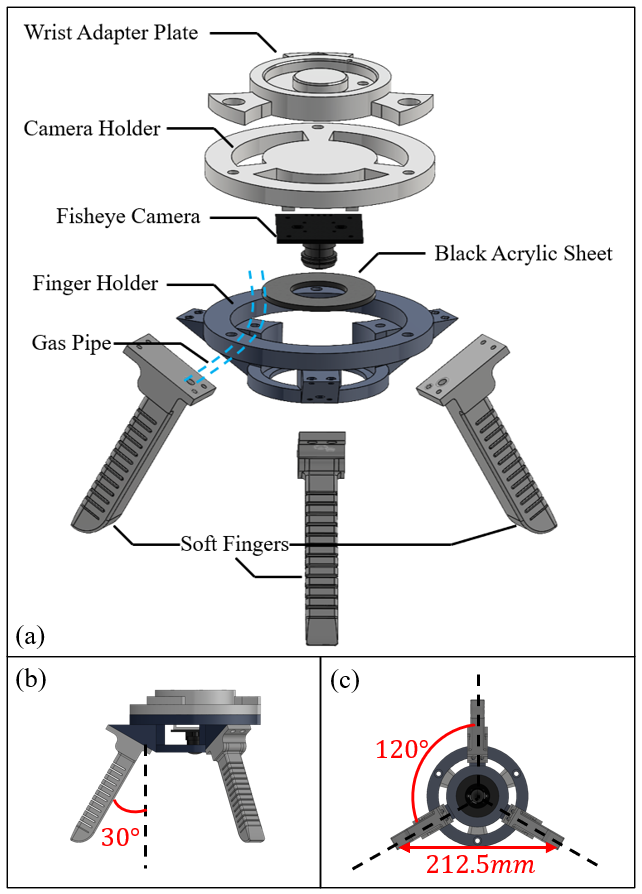}
    \caption{(a) The overview design of the gripper system. (b) Side view of the design. (c) Bottom view of the design. (To provide a clear view, the connecting screws are hidden to allow the principal elements to be easily visible.)}
    \label{structure}
\end{figure}

\begin{figure}[htbp]
    \centering
    \includegraphics[width=1.0\linewidth]{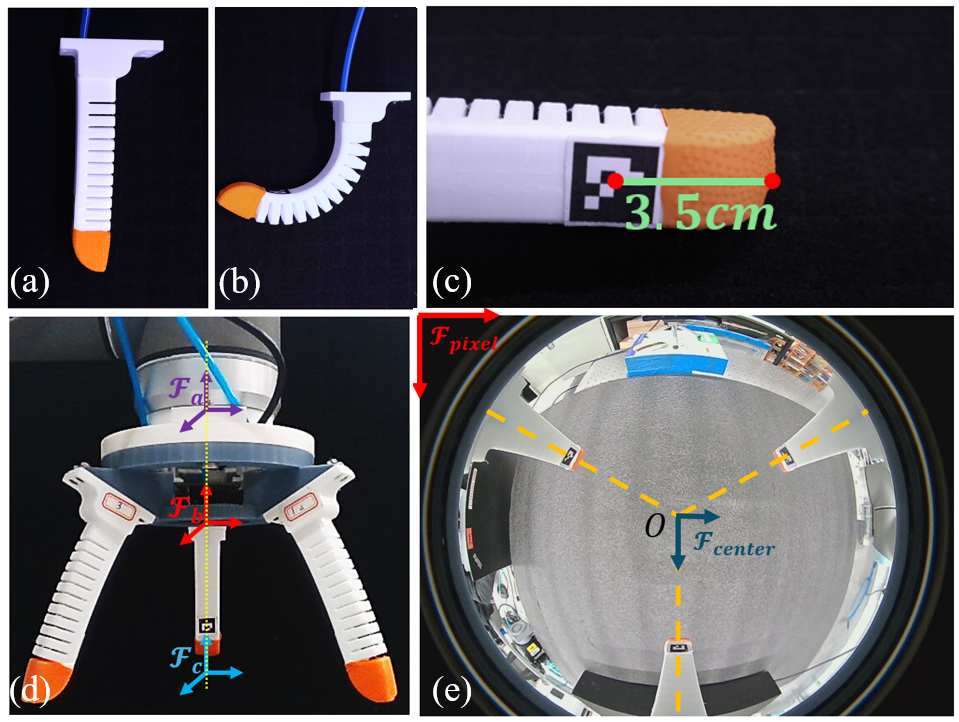}
    \caption{
    (a) Initial status of the soft finger.
    (b) Bending status of the soft finger.
    (c) The appearance of the fingertip with a rubber finger coat and Aruco marker.
    (d) The overview of the gripper system: $\mathcal{F}_{a}$ is the coordinate of the end-effector; $\mathcal{F}_{b}$ is the coordinate of the fisheye camera; $\mathcal{F}_{c}$ is the coordinate at the grasping center when the gripper is open.
    (e) The camera view: $\mathcal{F}_{pixel}$ is the coordinate of the frame; $\mathcal{F}_{center}$ is the coordinate of the grasping center reflected in the frame. }
    \label{fingers}
\end{figure}

The design of the gripper is shown in Fig. \ref{structure}(a), which highlights its key components for clarity. 
The gripper is supported by three necessary components: the finger holder, camera holder, and wrist adapter plate. 
The design follows modular principles, permitting each part of the gripper to be easily replaced without influencing other modules.
The supporting structure is 3D-printed by PLA, with excellent rigidity and durability, ensuring the gripper's robust performance. 
The gripper employs three pneumatically actuated pliant fingers, controlled by SMC ITV 0030-2S. 
Each finger can transform from a default straight state (Fig. \ref{fingers}(a)) to a bent grasping state (Fig. \ref{fingers}(b)) via pressurized actuation. 
The intake pipe is shown in blue airflow lines in Fig. \ref{structure}(a). 
Finger material selection balances shape conformity and proper payload. 
We fabricate fingers from TPU95A, which is a soft material but has a similar hardness to a skateboard wheel. 
In addition, rubber coatings shown in Fig. \ref{fingers}(c) are located at the fingertips, which decreases the possibility of slipping during dynamic manipulation. 
The gripper has a triangularly symmetric geometry: each finger is angled $30^{\circ}$ from the vertical axis (Fig. \ref{structure}(b)), with a $120^{\circ}$ angular separation between each two neighboring fingers (Fig. \ref{structure}(c)). 
These configurations lead to a large workspace for the gripper, where the distance between fingertips is $212.5 mm$, as shown in Fig. \ref{structure}(c).

The fisheye camera module has a wide range of $180^{\circ}$ to monitor both fingertip kinematics and target objects within a single frame. 
To resolve unclear soft finger motion tracking, Aruco markers are attached to the fingers at a distance of 3.5 mm from the fingertip, which reflects the deformation to marker displacement. 
To ensure a stable external optical environment, a concentric black acrylic sheet with an inner diameter of 32 mm and an outer diameter of 64 mm is used. 
Crucially, all the modules align the center with the gripper grasping center to eliminate complicated coordinate transformations during visual servo control, which is demonstrated in Fig. \ref{fingers}(d).

\subsection{Integrated Coordinate System}
\label{sec3_2}
To verify that the gripper's unified coordinate system can effectively reduce the computational cost, let $\mathcal{F}_{base}$, $\mathcal{F}_a$, $\mathcal{F}_b$, and $\mathcal{F}_c$ denote the robot base frame, end-effector frame, camera frame, and grasping center frame. 
The predefined transformation $T_{base}^{a}$ is related to the D-H parameters of robot arms. 
Meanwhile, $T_{a}^{b}$ and $T_{a}^{c}$ are also certain with the constraints of the gripper structure. 
The fisheye camera's intrinsic parameters establish a pixel-to-physical mapping to enable the centroid pixel frame $\mathcal{F}_{center}$ as the projection ($T^{c}_{center}$) of the 3D frame $\mathcal{F}_c$ through a fixed transformation between $\mathcal{F}_{center}$ and $\mathcal{F}_{pixel}$ ($T_{pixel}^{center}$), where $\mathcal{F}_{pixel}$ represents the image frame. 
Thus, we can establish the transformation from the robot base frame to the camera frame and grasping center frame:
\begin{equation}
    T_{base}^{b} = T_{base}^{a}T_{a}^{b}
\end{equation}
and
\begin{equation}
    T_{base}^{c} = T_{base}^{b}T_{b}^{c}=T_{base}^{b}T_{b}^{pixel}T_{pixel}^{center}T_{center}^{c}
\end{equation}
where $T_{b}^{pixel}$ is defined as the transformation from the camera frame to the image frame, which is fixed and dependent on the structure of the camera. In conclusion, by ascertaining $T_{pixel}^{center}$ offline and leveraging the components of the gripper's axis-aligned, the camera and grasping center can share the coordinate system. Above all, a direct connection and transformation among the robot, camera, and gripper is established, which provides the basic prerequisite for later implementation.

\section{Vision-based Collision Sensing Mechanism}
\label{sec4}

\subsection{Collision Analysis for Grasping Circular Objects}
\label{sec4_1}

\begin{figure}[h]
    \centering
    \includegraphics[width=1\linewidth]{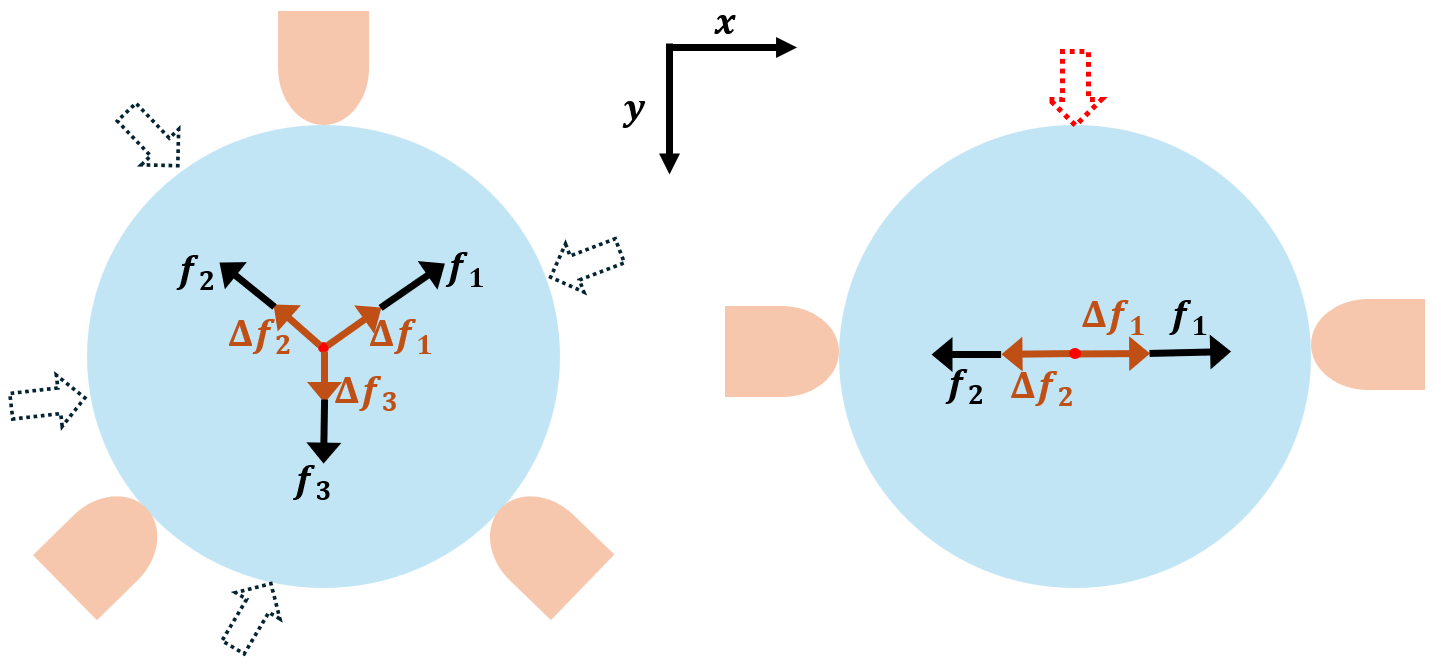}
    \caption{Comparison of three-fingered grasping model and two-fingered grasping model for circular objects. The former one can be more stable to burden multiple directional forces, while the latter one is always disturbed by the perpendicular forces to the fingers.}
    \label{analysis}
\end{figure}

For circular objects, the forces contacted on the surface are evenly distributed. For the three-fingered grasping model, the external disturbance forces can be decomposed into vectors in three directions, which can depict the two-dimensional force variations. Assuming the external disturbance force is $f_d$, the formulation can be written as:
\begin{equation}
    f_d=\Delta{f_1}+\Delta{f_2}+\Delta{f_3}
\end{equation}
where the two-dimensional expressions of $\Delta{f_1}$, $\Delta{f_2}$, $\Delta{f_3}$ are $(\Delta{x_1},\Delta{y_1})$, $(\Delta{x_2},\Delta{y_2})$, and $(0,\Delta{y_3})$.
The decomposed disturbance force is:
\begin{equation}
    [f_d]_x=\Delta{x_1}+\Delta{x_2}
\end{equation}
\begin{equation}
    [f_d]_y=\Delta{y_1}+\Delta{y_2}+\Delta{y_3}
\end{equation}

In comparison, the changes of forces only exist in the horizontal direction in the two-fingered grasping model. When an external force acts on the object in random directions, there will always be a perpendicular direction of decomposed forces that cannot be eliminated:
\begin{equation}
    f_d \neq \Delta{f_1}+\Delta{f_2}
\end{equation}
Because the two-dimensional expressions of $\Delta{f_1}$ and $\Delta{f_2}$ are $(\Delta{x_1},0)$, and $(\Delta{x_2},0)$. It cannot resist disturbance in the y-direction. Therefore, to keep stability and resist collision for circular object grasping, a three-fingered grasping model is necessary.

\subsection{Depth-free Grasping Strategy}
\label{sec4_2}
\begin{figure*}[t]
    \centering
    \includegraphics[width=1.0\linewidth]{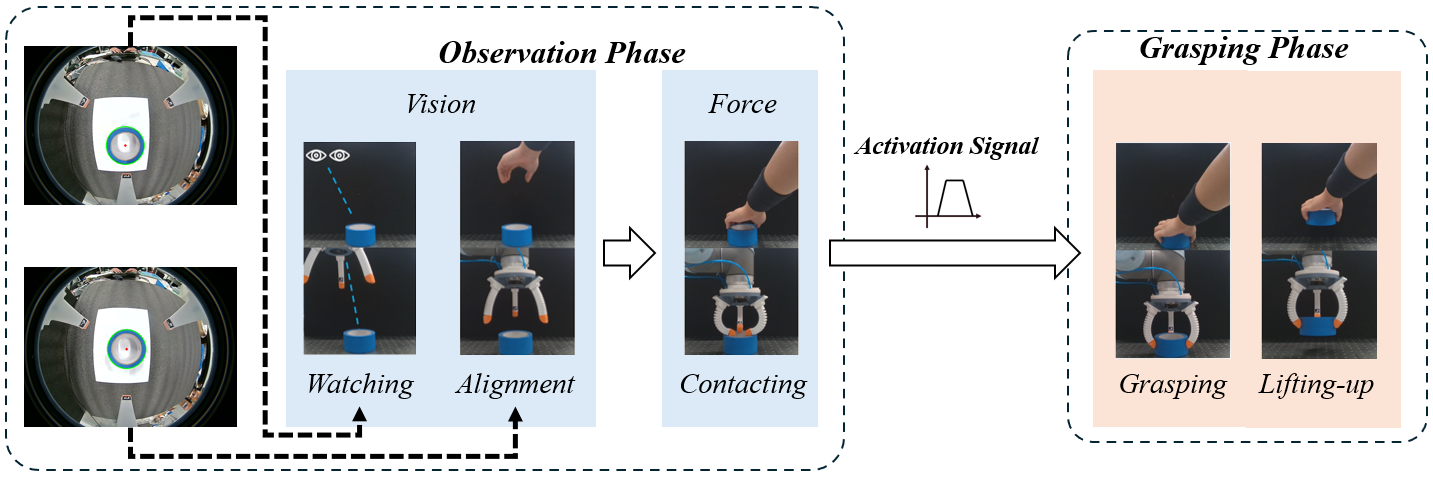}
    \caption{The entire pipeline of the grasping strategy inspired by human grasping. In observation phase, there are two components: vision and force. Visual information assists the actuator in aligning, and force information is used for localizing the object. Force also transfers a activation signal to the grasping phase and make the robot execute subsequent actions.}
    \label{H&G}
\end{figure*}

\begin{algorithm}[t]
  
  \SetAlgoLined
  
  \KwData{
  
  Gripper State: $g_{state}$, 
  
  Collision Vector: $\overrightarrow{C}$, 
  
  Collision Detection Constant: $threshold$, 
  
  Frame of Fisheye Camera: $F$
  
  Gripper Adjustment Constant: $z_g$}
  
  Initialization:\ 
  $g_{state}\xrightarrow{}open$\;
  \tcc{------ Observation Phase -----}
  $contour = HoughCircleDetection(F)$\;
  $center_x, center_y = PolygonCenter(contour)$\;
  
  Move horizontally to align the object:\
    $RobotCartesianMove([Dis(center_x), Dis(center_y), 0])$;
    
  Close the gripper and prepare to detect collision:\
  $g_{state}\xrightarrow{}close$\; 
  
  \While{$not \; ||\overrightarrow{C}||>=threshold$}{
    Move down to look for the object:\
    $RobotCartesianMove([0,0,-z])$;
  }
  \tcc{------- Grasping Phase -------}
  Move up and open the gripper:\
  $RobotCartesianMove([0,0,z_g])$;
  $g_{state}\xrightarrow{}open$;
  
  Move down and grasp the object:\
  $RobotCartesianMove([0,0,-z_g])$;
  $g_{state}\xrightarrow{}close$;

  Lift the object:\
  $RobotCartesianMove([0,0,z])$;
  \caption{Algorithm for Depth-Free Grasping}
\end{algorithm}

Traditional monocular grasping strategy such as \cite{9413197} depends on depth estimation from RGB frames. 
However, for frames from fisheye cameras, it is difficult to approach an accurate depth estimation due to the nonlinear radial distortion. 
Similarly, when humans grasp objects, they also cannot know the accurate numerical depth value.
Thus, humans typically use tactile information to assist in locating objects.
Inspired by human grasping behavior, we propose a novel grasping strategy that uses contact sensing to compensate for the lack of visual sensing.
Fig. \ref{H&G} displays the entire process of human grasping and the mimicked grasping strategy for the gripper. Initially, the humans observe and align the object by eyes while the gripper does so by the fisheye camera. Then humans will contact the object and obtain tactile feedback, which can also be conducted by the collision sensing for the gripper. Eventually, the hand and gripper execute grasping and lifting the object.

The principal algorithm framework is shown in \textbf{Algorithm 1}, comprising two major phases: the observation phase and the grasping phase. 
Preceding these phases, the gripper must align the center to the object to achieve optimal grasping quality.
The developed detector is based on the Hough circle detection algorithm, with the extraction of the position of the center.
Based on the inference of transformation between the robot frame and camera frame in Sec. \ref{sec3_2}, the robot can move in Cartesian space with the pixel-wise distance between the grasping center and the center of the object. 
In the observation phase, the gripper initiates in a \textit{close} state to expose Aruco markers to the camera. 
Then the robot continuously moves downward to try to approach the object until the collision occurs. 
The termination criterion for the process is when the scale of the collision vector $\overrightarrow{C}$ exceeds a predefined $threshold$, indicating the collision.
In this framework, the precise configurations of $\overrightarrow{C}$ are not required.
The trajectories of opening and closing for the soft fingers are not on the horizontal plane, which may lead to unexpected contact with the object. 
To avoid it, in the grasping phase, the robot will first move upward to reach the pre-grasp position and create sufficient space. 
Then it will move downward by the same distance to arrive at the grasp position.

\subsection{Extraction of Collision Information}
\label{sec4_3}
\begin{figure}[t]
    \centering
    \includegraphics[width=1.0\linewidth]{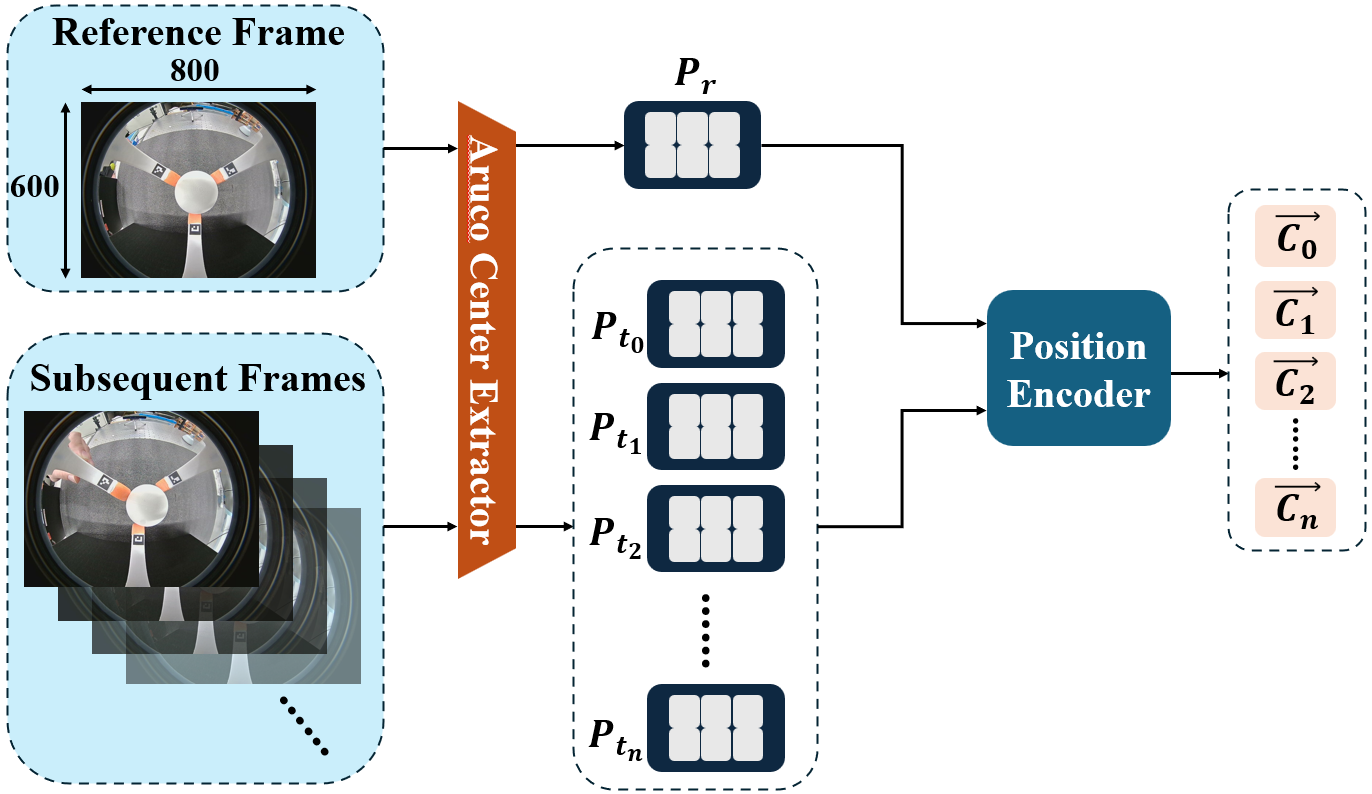}
    \caption{The complete pipeline involves extracting position matrices from captured frames and converting position information into collision vectors.}
    \label{pipeline}
\end{figure}

Collisions that occur during the post-grasping process have two typical formats, which are finger deformation and object displacement. 
In this section, we depict these two formats through a coordinated expression using a marker-based optical flow pipeline as shown in Fig. \ref{pipeline}. 
Initially, the fisheye camera captures the first frame as the reference frame ($800\:px\times600 \: px$). 
Subsequently, the frames are captured with the camera continuously at a frame rate of $20 \: fps$.
Via an \textit{Aruco Center Extractor}, the pixel-wise positions of the markers are extracted based on the Aruco module from OpenCV. 
Since all Aruco markers are positioned centrally within the fisheye camera frame, marker recognition remains unaffected by distortion.

The corresponding position matrix for the reference frame is $P_r$ in size of $3\times{2}$. 
For the subsequent frames, the position matrices follow the timeline of image capturing, which are annotated as $P_{t_0}$, $P_{t_1}$, $P_{t_2}$, ..., $P_{t_n}$. 
Furthermore, we developed a position encoder to integrate the position information. The encoder uses two position matrices as input to calculate its corresponding collision vector with the formulation: 
\begin{equation}
    (\overrightarrow{C_n})_{2\times1}=\sum_{i=0}^{2}{(P_{t_n}[i]-P_r[i])}
\end{equation}
where $\overrightarrow{C_n}\;(\Delta{y},\Delta{x})$ denotes the collision vector produced from the $n$ subsequent frame. To indicate the direction and the scale of the collision vector, we use polar coordinates to represent this vector: $\overrightarrow{C_n}\;(\Delta{r},\Delta{\theta})$, in which:
\begin{equation}
    \Delta{r}=||\overrightarrow{C_n}||=\sqrt{\Delta{y}^2+\Delta{x}^2}
\end{equation}
\begin{equation}
    \Delta{\theta}=\angle{\overrightarrow{C_n}}=\arctan(\frac{\Delta{x}}{\Delta{y}})
\end{equation}

\section{Experiments and Results}
\label{sec5}

\subsection{Grasping Tests}
\label{sec5_1}

\begin{figure}[t]
    \centering
    \includegraphics[width=1.0\linewidth]{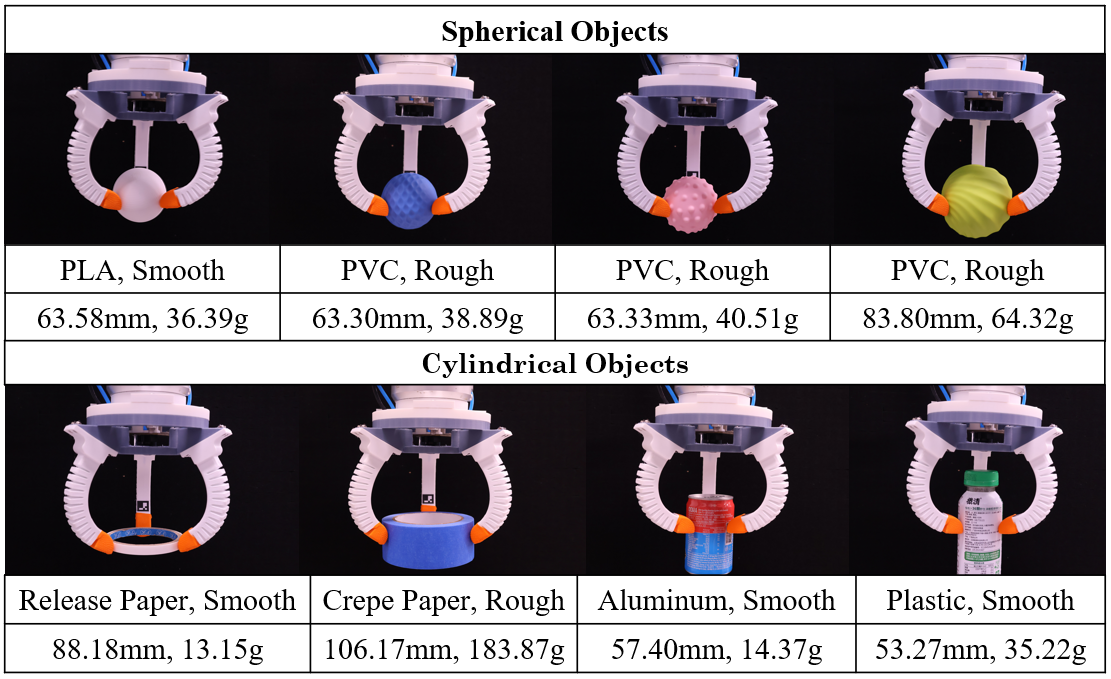}
    \caption{Circular objects used for grasping with different physical properties, including materials, smoothness, radius, and weights.}
    \label{Grasp}
\end{figure}

We conducted a series of grasping experiments to evaluate the gripper's capability to grasp everyday circular objects with diverse material properties, smoothness, radii, and weights. 
The principal categories of circular objects include both spherical items such as balls and cylindrical items such as bottles and tape rolls, as shown in Fig. \ref{Grasp}. 
The results demonstrated that our soft gripper can not only achieve precise and stable grasping of rigid circular objects but also successfully preserve the shape of soft deformable objects (e.g., soft PVC toy balls and empty bottles) during manipulation – a task that is typically challenging for rigid grippers or conventional two-finger grippers. 
Although the payload of the soft hand is not large, most small objects in everyday life can be easily picked up by this soft hand.

\subsection{Response Time Test}
\label{sec5_2}

\begin{figure}[h]
    \centering
    \includegraphics[width=0.8\linewidth]{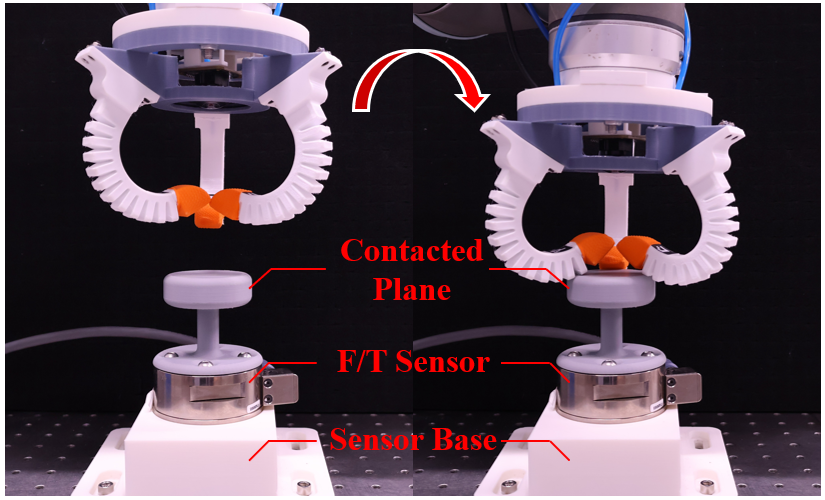}
    \caption{Experimental setup for contact response time test}
    \label{FT}
\end{figure}

To qualify the detection latency of the vision-based method, we employed the ATI Gamma Force and Torque (F/T) sensor as a high-frequency (2000 $Hz$ in $F_z$) reference. Since the frequency of the sensor exceeds the acquisition rate of the fisheye camera by a factor of 100, this setup establishes a ground-truth baseline, minimizing temporal aliasing. Fig. \ref{FT} demonstrates the experimental setup, where a 3D-printed cylinder-like object is mounted on the F/T sensor to provide a standard contact plane, and a sensor base is utilized to support the sensor.

\begin{figure}[h]
    \centering
    \includegraphics[width=1\linewidth]{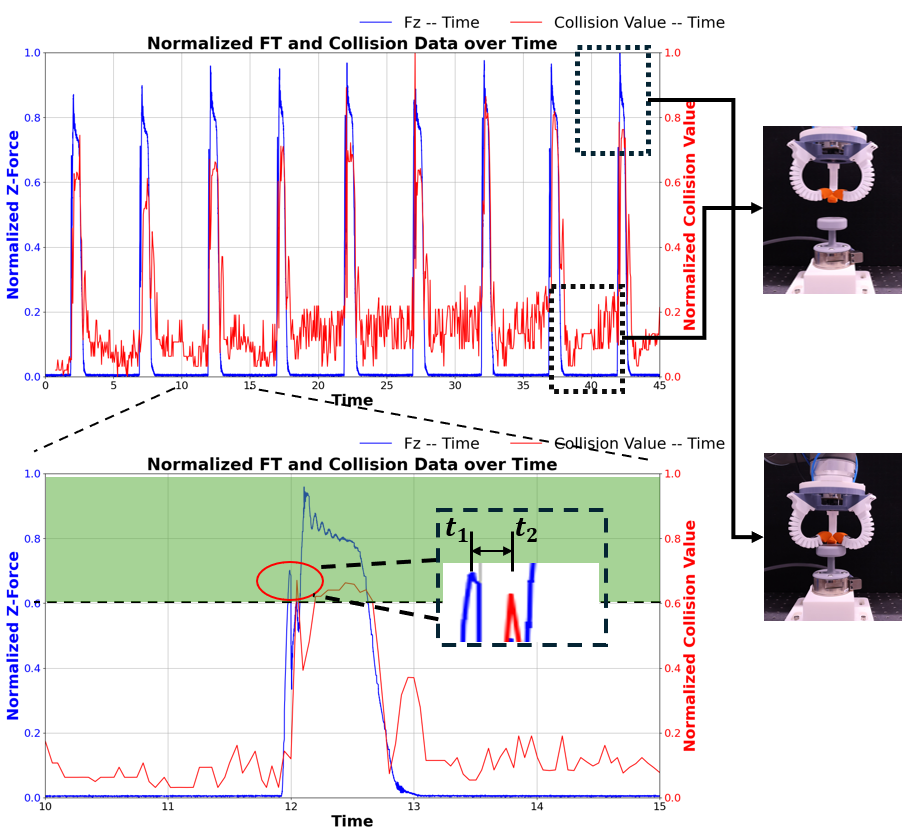}
    \caption{The collision sensing signals from the F/T sensor and camera with regular fluctuation. One period is zoomed in to visualize the annotation of response time in this work.}
    \label{RTresult}
\end{figure}

The gripper initiates in the closed position and moves downward at a speed of 0.5 $m/s$ to smash into the contact plane. After repeating the entire action sequence multiple times, a normalized result is displayed. To visualize the response time directly, a random period is selected and zoomed in. The force features collected from the experiments are normalized, with collision onset defined as the moment when the first data has a percentage greater than 60\% of peak force. This threshold can immunize the noise and keep the sensitivity. As shown in \ref{RTresult}(b), the system response time can be computed as:
\begin{equation}
    ART=\frac{1}{n}\sum_{i=1}^{n}(t_{2,i}-t_{1,i})
\end{equation}
where $t_{1,i}$ and $t_{2,i}$ denote the vision-based collision detection time and F/T sensing detection onset at the period $i$ respectively. After multiple trials, the average response time is 204.75 msec, which is even less than the human tactile reaction latency of 241 msec mentioned in \cite{kim2020visual}. This validates that this gripper system possesses real-time collision detection capabilities with only an RGB frame as feedback.

\subsection{Collision Information Visualization}
\label{sec5_3}

\begin{figure}[h]
    \centering
    \includegraphics[width=1\linewidth]{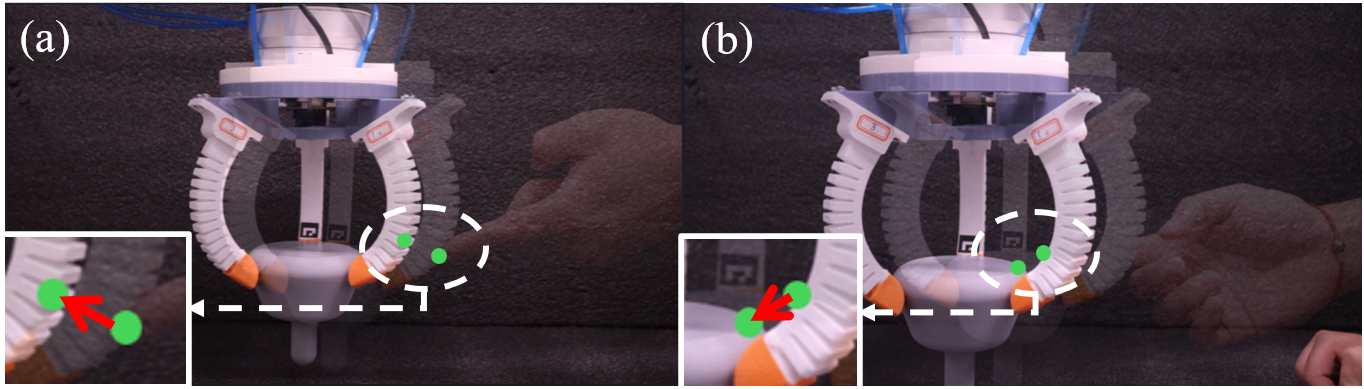}
    \caption{(a) The gripper perceives the collision on finger and dodges; (b) The gripper perceives the collision on grasped object and dodges.}
    \label{dodge}
\end{figure}

Some trials are conducted to illustrate the avoidance mechanism for the gripper. 
In most commercialized grippers, systematic anomaly detection is typically based on abnormal variations in current. 
However, their mechanism often causes the system to shut down after a critical event, preventing the robot from quickly escaping a dangerous environment. 
As depicted in Fig. \ref{dodge}, the robot can move in Cartesian coordinates upon sensing collisions from different directions and magnitudes to assist the robot in leaving. 
The dodging direction aligns precisely with the collision force applied to the finger or object.

\begin{figure*}[htbp]
    \centering
    \includegraphics[width=\textwidth]{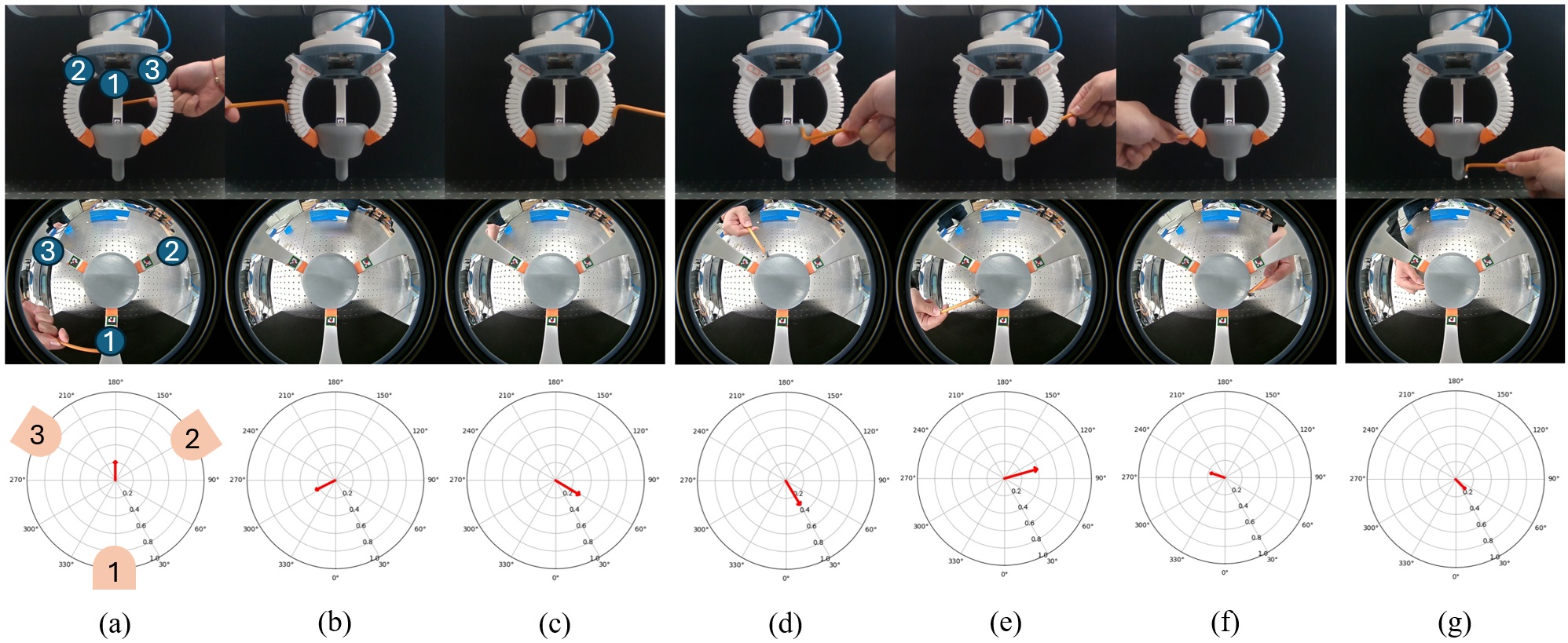}
    \caption{(a)-(c): Collision test on soft fingers. (d)-(f): Collision test on grasped object. (g): Collision test on an unseen part of the grasped object. The first row demonstrates the overall view of the gripper; the second row displays the camera view; the third row shows the polar plot. }
    \label{all}
\end{figure*}

To verify the reliability of the collision sensing, we conducted experiments using a 3D-printed pumpkin-like object designed to emulate real-world grasping complexities. 
A series of demonstrations is displayed in Fig. \ref{all}. 
Specifically, the serial numbers of all three fingers are annotated in Fig. \ref{all}(a). 
In the polar plot, the respective fingers are distributed at $0^{\circ}$, $120^{\circ}$, and $240^{\circ}$. Collisions are induced manually using an L-shaped hexagon wrench.

From Fig. \ref{all}(a) to Fig. \ref{all}(c), there are three trials regarding testing the collision direction on the three fingers respectively. 
The outcomes indicate accurate directional sensing for external disturbances. 
The mathematical representations of these are:
\begin{equation}
\angle{\overrightarrow{C_{f1}}}=180^{\circ}, \angle{\overrightarrow{C_{f2}}}=300^{\circ}, \angle{\overrightarrow{C_{f3}}}=60^{\circ}
\end{equation}
where $\overrightarrow{C_{f1}}$, $\overrightarrow{C_{f2}}$, and $\overrightarrow{C_{f3}}$ are the collision vectors of the three fingers. All vectors and their corresponding fingers are distributed $180^\circ$.
From Fig. \ref{all}(d) to Fig. \ref{all}(f), we applied three random contact forces to the grasped object and the results show the correct sensing direction for all of the collisions. 
In Fig. \ref{all}(g), we conducted a collision test on the unseen tiny head to verify that the system can also detect collisions that occur on the invisible part. 
Compared with the first row and the second row, the fisheye cameras cannot capture all the tapping actions from humans like Fig. \ref{all}(b) and Fig. \ref{all}(c), so it is impossible to use cameras to accurately predict the occurrence of collisions. Therefore, it is meaningful for the vision-based collision sensing mechanism to respond and react to collision efficiently.

\section{Conclusion}
In this paper, we introduced a collision sensing module for circular object grasping and verified its stability with a soft gripper system. We first illustrated the reliability of the gripper system with two experiments. The first experiment demonstrated its capability to grasp various circular objects, while the second experiment indicated the system is sensitive enough to perceive external disturbances. Additionally, a collision-rich grasping strategy and a collision detection mechanism were presented. We showed the pipeline of the grasping strategy and the inferences of the collision vectors, respectively. Finally, a collision information visualization was employed to denote the precise result of collision detection. In future work, we may focus on improving the gripper's configuration to increase grasping force, enhancing the visual algorithms for more precise grasping, and generalizing to the manipulation of deformable objects. 

\bibliographystyle{ieeetr} 
\bibliography{reference}

\end{document}